\documentclass[conference]{IEEEtran}
\IEEEoverridecommandlockouts
\usepackage{cite}
\usepackage{amsmath,amssymb,amsfonts}
\usepackage{algorithmic}
\usepackage{graphicx}
\usepackage{textcomp}
\usepackage[table]{xcolor}
\usepackage{url}
\usepackage{subcaption}
\usepackage{amssymb}
\usepackage{pifont}
\newcommand{\xmark}{\ding{55}}%
\usepackage{booktabs}
\usepackage[table]{xcolor}
\usepackage{wrapfig,lipsum,booktabs}

\usepackage{multirow}
\usepackage{lipsum}
\usepackage{hyperref}
\usepackage{cleveref}
\usepackage[colorinlistoftodos]{todonotes}
\newcommand\copyrighttext{%
    \small \begin{center} \color{gray} \textcopyright\,2025 IEEE.  Personal use of this material is permitted.  Permission from IEEE must be obtained for all other uses, in any current or future media, including reprinting/republishing this material for advertising or promotional purposes, creating new collective works, for resale or redistribution to servers or lists, or reuse of any copyrighted component of this work in other works. \end{center}}

\crefname{figure}{Fig.}{Figs.}

\definecolor{darkgreen}{rgb}{0.0, 0.39, 0.0}

\begin{document}

\title{\vspace{-2em} \copyrighttext \vspace{1em} \Huge CUBIC: Concept Embeddings for Unsupervised Bias Identification using VLMs}

\author{
\IEEEauthorblockN{%
David Méndez\IEEEauthorrefmark{1}\thanks{Corresponding author: davidmendez@ugr.es},
Gianpaolo Bontempo\IEEEauthorrefmark{2},
Elisa Ficarra\IEEEauthorrefmark{2},
Roberto Confalonieri\IEEEauthorrefmark{3}, and
Natalia Díaz-Rodríguez\IEEEauthorrefmark{1}
}
\IEEEauthorblockA{%
\IEEEauthorrefmark{1}\textit{Dept. of Computer Science and Artificial Intelligence, DaSCI Institute, University of Granada}, Granada, Spain\\
\IEEEauthorrefmark{2}\textit{Dept. of Engineering "Enzo Ferrari", University of Modena and Reggio Emilia}, Modena, Italy\\
\IEEEauthorrefmark{3}\textit{Dept. of Mathematics 'Tullio Levi-Civita', University of Padova}, Padova, Italy
}
}

\maketitle

\begin{abstract}
Deep vision models often rely on biases learned from spurious correlations in datasets. To identify these biases, methods that interpret high-level, human-understandable concepts are more effective than those relying primarily on low-level features like heatmaps. A major challenge for these concept-based methods is the lack of image annotations indicating potentially bias-inducing concepts, since creating such annotations requires detailed labeling for each dataset and concept, which is highly labor-intensive. 
We present CUBIC (\textbf{C}oncept embeddings for \textbf{U}nsupervised \textbf{B}ias \textbf{I}dentifi\textbf{C}ation), a novel method that automatically discovers interpretable concepts that may bias classifier behavior. Unlike existing approaches, CUBIC does not rely on predefined bias candidates or examples of model failures tied to specific biases, as these are not always available in the data. Instead, it utilizes image-text latent space and linear classifier probes to examine how the latent representation of a superclass label---shared by all instances in the dataset---is influenced by the presence of a concept.
By measuring these shifts against the normal vector to the classifier's decision boundary, CUBIC identifies concepts that significantly influence model predictions. Our experiments demonstrate that CUBIC effectively uncovers previously unknown biases using Vision-Language Models (VLMs) without requiring the samples in the dataset where the classifier underperforms or prior knowledge of potential biases. 
\end{abstract}


\section{Introduction}\label{sec:introduction}

Computer vision has transformed various industries by equipping machines with the ability to perform complex visual tasks traditionally requiring human intelligence. From medical diagnosis and autonomous driving to facial recognition and manufacturing quality control, these systems have achieved remarkable advancements. However, as these systems become increasingly integrated into critical decision-making processes, a significant concern has emerged: deep learning vision models can exhibit bias, leading to unintended outcomes such as inequitable decision-making, discrimination, or the amplification of existing societal disparities \cite{mehrabi2021survey}. These biases can manifest in various forms, from gender and racial prejudices to socioeconomic discrimination, potentially affecting individuals who interact with these systems daily.

The critical nature of this problem has spawned numerous attempts to understand and visualize how these models make decisions. Historically, solutions highlighting specific parts of the image have been the most popular ones \cite{selvaraju2017grad, simonyan2014deep, ribeiro2016should, sundararajan2017axiomatic}. These approaches, commonly known as attribution or saliency methods, attempt to identify which regions of an input image most strongly influence a model's predictions. However, these heatmap-based visualization methods face several fundamental limitations. Not only do they suffer from faithfulness issues \cite{adebayo2018sanity}, where the highlighted regions may not truly reflect the model's decision-making process, but they are also vulnerable to manipulation \cite{anders2020fairwashing}, allowing malicious actors to create misleading explanations. Perhaps most importantly, since these methods work with low-level features such as pixel intensities or activation patterns, they fail to link model decisions to human-understandable concepts, making it difficult for practitioners to identify and address systemic biases.

Therefore, to identify biases in vision models, which often arise from spurious correlations in datasets, it is preferable to use methods that convey high-level, human-understandable concepts rather than relying on visualizations \cite{poeta2023concept}. For instance, instead of highlighting pixels in an image, a more helpful approach would identify that a model is biased towards making predictions based on background scenery rather than the actual object of interest. However, the primary challenge in this bias identification approach is the absence of image annotations that specify potentially bias-inducing concepts, as creating these annotations would be labor-intensive, requiring detailed labeling for each dataset and each concept.

\begin{figure*}[!t]
    \centering
    \begin{subfigure}[t]{0.3\textwidth}
        \centering
        \includegraphics[width=\textwidth]{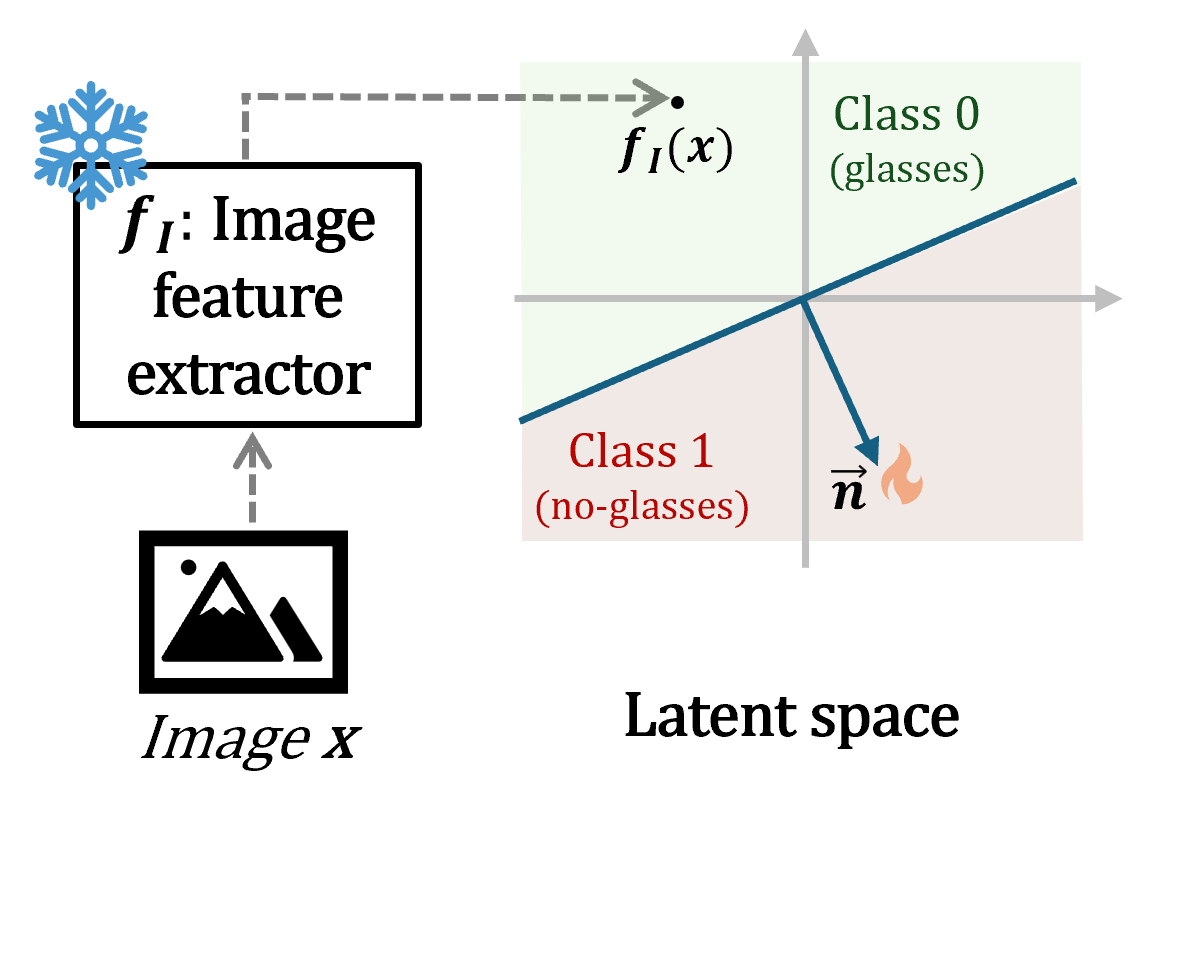}
        \caption{Linear probe classifier. Example: binary classifier for wearing vs not glasses.}
        \label{fig:linear_probe}
    \end{subfigure}
    \hspace{0.001\textwidth}
    \begin{subfigure}[t]{0.35\textwidth}
        \centering
\includegraphics[width=\textwidth]{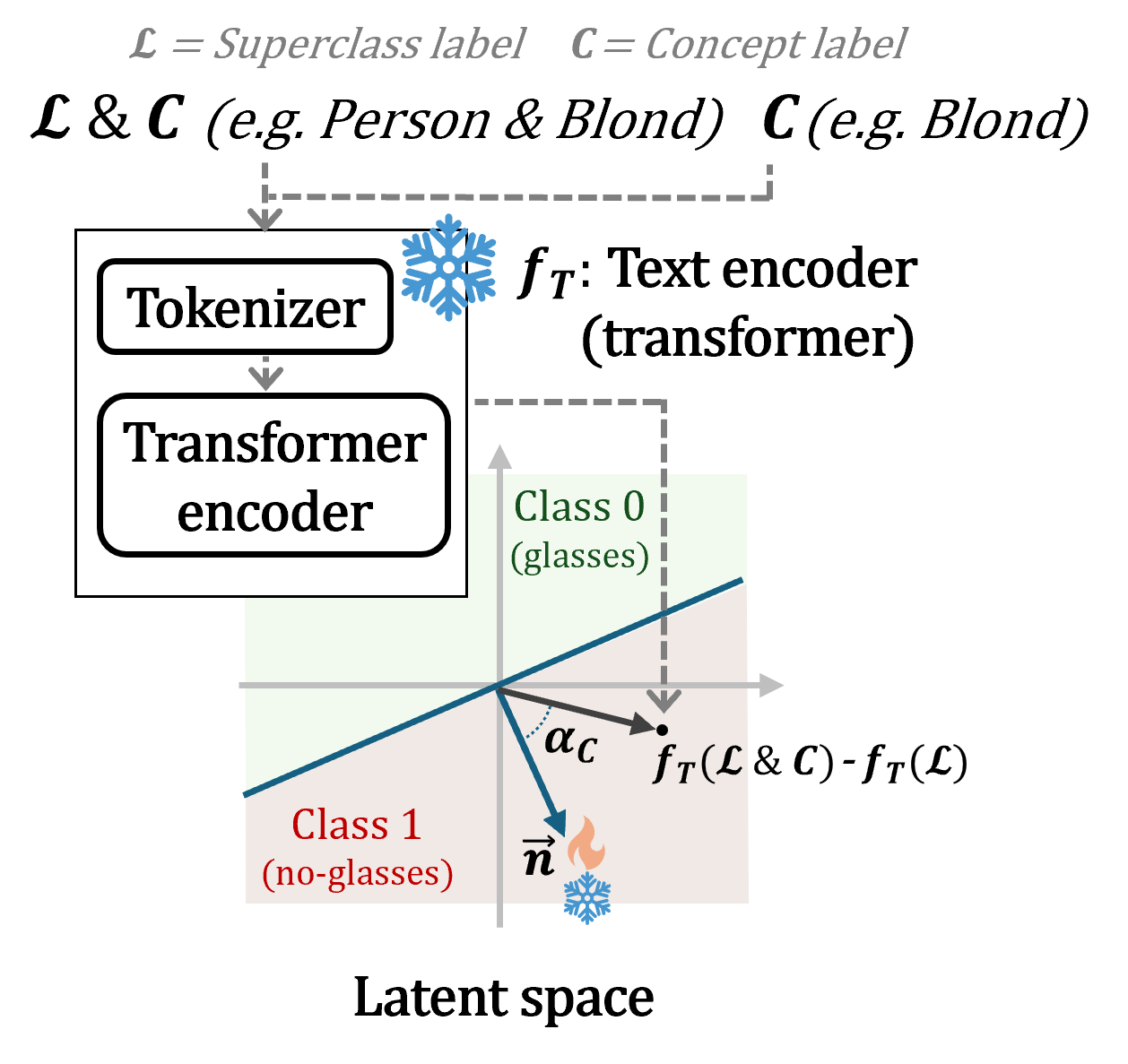}
        \caption{Angle $\alpha_{C}$ between normal vector $\vec{n}$ and concept-driven shift in superclass $\mathcal{L}$ embedding.} 
        \label{fig:angle_between_concept_and_hyperplane}
    \end{subfigure}
        \hspace{0.001\textwidth}
    \begin{subfigure}[t]{0.25\textwidth}
        \centering
\includegraphics[width=\textwidth]{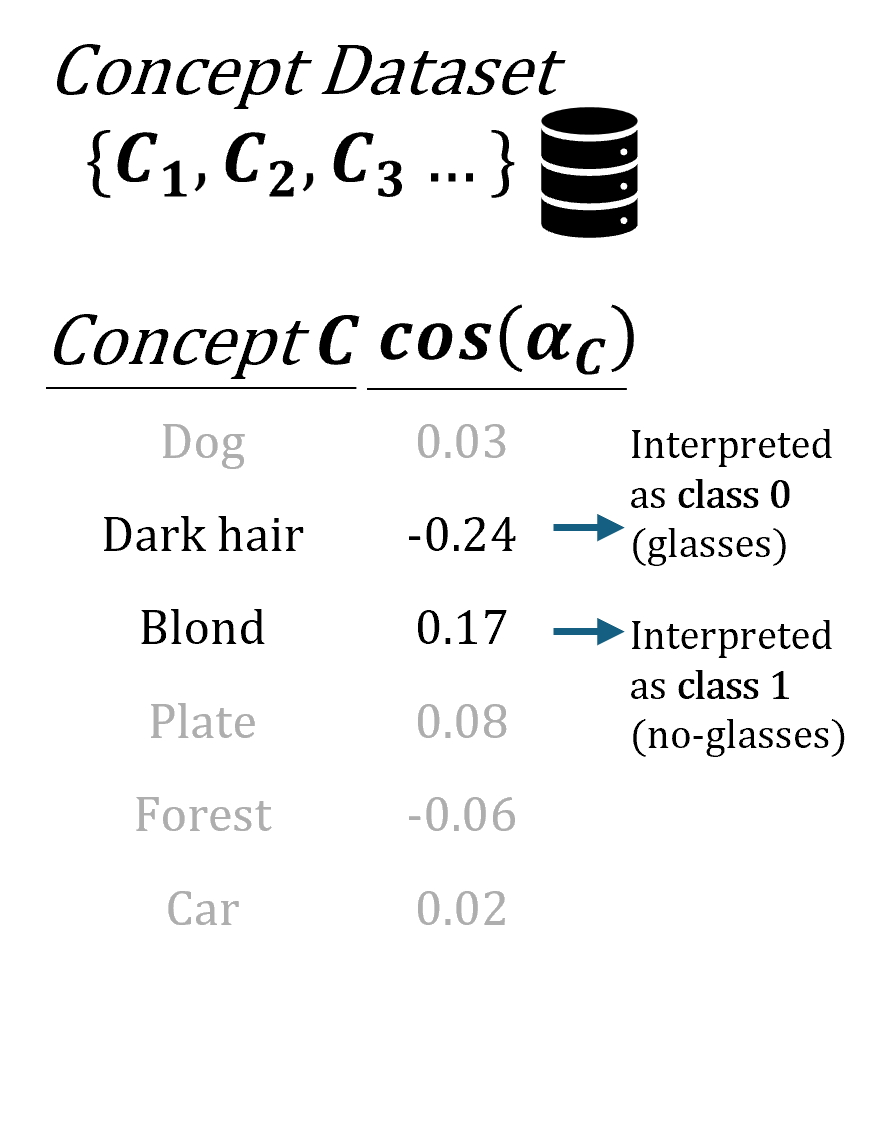}
        \caption{Bias-inducing concept $C$ identification.}
        \label{fig:selection_of_concepts}
    \end{subfigure}
    \caption{CUBIC methodology illustrated.  In (\subref{fig:linear_probe}), a linear probe classifier is constructed by training a linear SVM on the features provided by a frozen image encoder. In (\subref{fig:angle_between_concept_and_hyperplane}), the cosine of the angle $\alpha_{C}$ between the vector normal to the SVM hyperplane, $\vec{n}$, and the concept-driven shift in superclass embedding $f_{T}(\mathcal{L} \land C) - f_{T}(\mathcal{L})$ is calculated. Here, $f_{T}(\mathcal{L})$ represents the embedding of a superclass label common to all images (e.g., \textit{Person} in the CelebA dataset \cite{liu2015faceattributes}, \textit{Bird} in the Waterbirds dataset \cite{sagawadistributionally}). On the other hand, $\mathcal{L} \land C$ represents a prompt combining concept $C$ and its superclass label $\mathcal{L}$ (e.g., \textit{Person, Blond}), and $f_{T}(\mathcal{L} \land C)$ its embedding in the latent space. 
    In (\subref{fig:selection_of_concepts}), $\cos\alpha_{C}$ indicates the magnitude and the class to which
    concept $C$ biases the model.
    If $\cos\alpha_{C} > 0$, the concept-driven shift of the superclass embedding $f_{T}(\mathcal{L} \land C) - f_{T}(\mathcal{L})$ points toward the class $\bf{1}$ side of the hyperplane. This means concept $C$ \textit{pushes} the superclass embedding in the direction
    of class $\bf{1}$ (no-glasses). The opposite occurs when $\cos\alpha_{C} < 0$. Had we taken $\vec{n}$ towards class $\bf{0}$, $\cos\alpha_{C} > 0$ would indicate the concept is pushing toward class $\bf{0}$. 
    }
    \label{fig:visual_abstract}
    \frenchspacing 
\end{figure*}

Recent methods have emerged aimed at identifying bias using human-understandable concepts in an unsupervised manner \cite{eyuboglu2022domino, jain2023distilling, kim2024keyword, rezaei2024prime, zhang2023diagnosing}. These approaches represent significant progress in automating bias detection without requiring extensive manual annotation. Nonetheless, they face important limitations. Indeed, many of these methods rely on detecting performance degradation across specific subpopulations within the dataset \cite{kim2024keyword, rezaei2024prime, jain2023distilling, eyuboglu2022domino, wiles2022bugs}, which may not be adequately represented in available data samples. This dependence on subpopulation performance can be particularly problematic when dealing with underrepresented groups or edge cases. Other approaches are limited to detecting bias from a predefined set of possible biases \cite{zhang2023diagnosing}, potentially missing novel or unexpected forms of bias that weren't anticipated during system design. For this reason, we propose \textbf{C}oncept embeddings for \textbf{U}nsupervised \textbf{B}ias \textbf{I}dentifi\textbf{C}ation CUBIC
\footnote{Code available at \url{https://github.com/david-mnd/CUBIC}.}
(illustrated at \Cref{fig:visual_abstract}), a novel solution to detect concept-induced bias on a linear classifier probe fine-tuned on top of a visual-language model (VLM). 
Rather than focusing on performance metrics or predefined bias categories, CUBIC measures how the latent representation of a superclass label — shared by all images in the dataset — shifts in response to the presence of a specific concept. This approach enables us to understand a concept's effect on the linear probe model in a specific classification task.

\textbf{Contribution}.
Our method offers several key advantages over existing approaches: 
\begin{itemize}
    \item CUBIC can identify bias in a linear classifier probe without requiring access to failure cases where the classifier underperforms. This sets it apart from most existing bias identification methods, which typically rely on detecting performance disparities across different subgroups. By analyzing representation shifts rather than performance metrics, CUBIC can potentially identify biases before they manifest as observable failures.

    \item The system automatically identifies those concepts most associated with bias, even without requiring a restricted list of candidate concepts to be tested for bias induction. This capability allows CUBIC to discover unexpected or novel sources of bias that might be missed by approaches that rely on predefined bias categories. 

\end{itemize}

\section{Related Work}\label{sec:relwork}

\noindent
\textbf{Challenges in Bias detection}.
Solutions that highlight specific regions of an image where a model focuses, such as saliency maps \cite{selvaraju2017grad, simonyan2014deep, ribeiro2016should, sundararajan2017axiomatic}, have proven effective in detecting model biases. 
For instance, using saliency maps, a study \cite{degrave2021ai} revealed that classifiers trained to detect COVID-19 cases from chest X-rays focused on spurious signals, such as text markers or imaging artifacts, rather than medically relevant evidence. This highlights how visualization tools can uncover unintended biases in model behavior. However, these methods often require human intervention to interpret visualizations, suffer from faithfulness issues \cite{adebayo2018sanity}, and are vulnerable to manipulation \cite{anders2020fairwashing}. Additionally, while these techniques indicate where the model is focusing, they fail to explain the concept in the highlighting region the model focuses on \cite{Stammer_2021_CVPR}, leaving a critical gap in understanding the underlying reasoning behind the model's predictions.
Therefore, using human-understandable, concept-based methods for bias detection in deep vision models is more effective. However, manually annotating datasets to identify bias-inducing concepts is prohibitively
time-consuming.

\smallskip
\noindent
\textbf{Bias detection from performance degradation.}
With the advent of Vision-Language Model (VLM) encoders, several approaches have emerged to identify bias without requiring concept annotations. This is done by leveraging the shared latent space of image-text representations. They automatically assign concepts to a group of images where the classifier struggles. For instance, DOMINO \cite{eyuboglu2022domino} uses Gaussian Mixture Models (GMM) in the vision-language representation space to identify regions where model performance drops, associating these regions with natural language descriptions. Similarly, Distilling-Failures \cite{jain2023distilling} employs an SVM on the VLM latent space to distinguish between correctly and incorrectly predicted images. They retrieve captions for samples farthest from the hyperplane in the direction of wrong labels, labeling them as "hard samples" to reveal spurious correlations. Other works automate bias discovery using captioning or tagging methods on images where the model fails \cite{kim2024keyword, wiles2022bugs, rezaei2024prime}. These methods post-process extracted concepts using similarity scores in a VLM latent space \cite{kim2024keyword}, refine descriptions via generative models \cite{wiles2022bugs}, or search for concept combinations causing significant performance drops \cite{rezaei2024prime}.

In particular, \textit{Bias2Text} (B2T) \cite{kim2024keyword} does not require a restricted list of candidate concepts for bias induction and is a powerful approach for identifying concepts linked to classification biases. B2T begins by storing captions for all images using an image captioning method. Then, the YAKE keyword extraction method \cite{campos2020yake} is applied to captions of misclassified images to extract a set of concepts associated with classifier failures. Finally, a CLIP score is used to quantify how close a concept is to the misclassified images compared to the correctly classified images, measuring the concept's relationship to the errors. 
Even though previous methods provide valuable insights regarding bias detection, they all rely on samples where the classifier underperforms, which may not always be available. Since these methods solely rely on identifying errors within the dataset, they are limited to uncovering explicitly represented biases. In contrast, our approach overcomes this data limitation (\Cref{table:bias_detection_methods}) by moving the bias discovery completely to the latent space, enabling the identification of potential biases beyond the constraints of the available dataset.

\noindent
\textbf{Bias detection beyond performance degradation.}
 A limitation of the previous methods is their reliance on the presence of misclassified images linked to the bias. Indeed, since validation and test sets are typically drawn from the same distribution as training data, misclassified images linked to the bias may not be present. In the literature, there is a lack of solutions capable of detecting bias without relying on misclassified samples to reveal such bias. To the best of our knowledge, only \textit{DrML}~\cite{zhang2023diagnosing} addresses this issue. Also \textit{DrML} shows that linear classifiers built on a VLM latent space possess the property of cross-modal transferability \cite{zhang2023diagnosing}, which allows a classifier trained on latent image representations to accurately process latent text representations as well. \textit{DrML} uses textual fine-grained classes from an image dataset to feed the linear classifier on latent space. For example, in the Waterbirds dataset \cite{sagawadistributionally}, which includes two bird categories (\textit{Waterbirds} and \textit{Landbirds}), the fine-grained classes are the specific bird species. 
\textit{DRML} then calculates an influence score, quantifying the average change in the classifier's predicted probability when a text concept is introduced alongside the fine-grained class.
However, \textit{DRML} relies on a predefined set of bias candidate concepts, requiring prior knowledge of potential biases. On the contrary, our method builds a classification task-independent concept dataset to detect bias without predefined candidates.

\begin{table}[!t]
\caption{
Comparison of requirements for bias identification methods (misclassified samples linked to the bias and bias-inducing concepts candidates).}
\centering
\begin{tabular}{ccc}
\toprule
\textbf{
Bias Disc. Method} & \textbf{No Misclass. samples} & \textbf{No concept candidates} \\ 
\midrule
B2T \cite{kim2024keyword}            &    \textcolor{red}{\xmark}                                 & \textcolor{darkgreen}\checkmark                          \\
DrML   \cite{zhang2023diagnosing}                      & \textcolor{darkgreen}\checkmark                                       & \textcolor{red}\xmark                         \\
\textbf{CUBIC} (ours)           & \textcolor{darkgreen}\checkmark                                       & \textcolor{darkgreen}\checkmark                          \\
\bottomrule
\end{tabular}
\vspace{-10pt}
\label{table:bias_detection_methods}
\end{table}

\section{Methodology}\label{sec:background}

This section presents CUBIC, a novel methodology for unsupervised bias identification in Vision-Language Models (VLMs) (see \Cref{fig:visual_abstract}). At its core, CUBIC employs a quantitative bias scoring mechanism to systematically identify and extract the most significant bias-inducing concepts from a curated concept dataset. The methodology comprises four key components:
\begin{enumerate}
\item[\textit{A)}] \textit{Finetune a linear classifier probe:} We construct a linear probe by training a linear classifier on feature representations extracted from a frozen VLM image encoder. 
\item[\textit{B)}] \textit{Create the concept-based dataset:} We create a task-agnostic dataset from which biasing concepts will be extracted.
\item[\textit{C)}] \textit{Compute the CUBIC bias score:} This score quantifies the degree to which each concept is assumed by the model as an evidence of a class to be predicted.
\item[\textit{D)}] \textit{Identify bias-inducing concepts:} We leverage the computed CUBIC bias scores to identify the most significant bias-inducing concepts.
\end{enumerate}

\subsection{Finetune a linear classifier probe.}
\noindent Training a fully-connected layer on top of a frozen visual foundation model feature extractor is an efficient way to build a high-performing classifier. This technique, known as \textit{linear probing} \cite{alain2017understanding}, is commonly used to evaluate visual foundation models feature extractors \cite{he2020momentum}. Recent research has shown that feature extractors trained in weakly-supervised \cite{radford2021learning, cherti2023reproducible, fang2023eva} or self-supervised \cite{oquab2023dinov2, zhou2021ibot, caron2021emerging} settings can be used to build linear probes with impressive performance\cite{radford2021learning}.

\noindent
\textbf{VLMs and Cross-modal transferability.}
VLM feature extractors, such as CLIP or ALIGN \cite{jia2021scaling}, consist of both an image and a text encoder. Both encoders produce embeddings in the same latent space, ensuring that text-image pairs have similar representations. 
 Although VLM feature extractors, like CLIP, encode images and texts into the same latent space, a modality gap exists \cite{liang2022mind} that causes image and text embeddings to occupy different regions of this space. Nonetheless, \cite{zhang2023diagnosing} shows that linear classifiers with no summing terms acting on top of the latent space (see Eq. \eqref{eq:linear_probe}) can produce similar outputs when they are fed an image or its text description. The authors of \cite{zhang2023diagnosing} call this property \textit{cross-modal transferability}. Due to cross-modal transferability, we can analyze how a concept present in images impacts the linear classifier probe by examining the embedding of its textual description.

\noindent
\textbf{Notation.} 
Given the input image space \(X\), let \(f_I: X \rightarrow Z\) be an image encoder that maps \(X\) to the latent space \(Z\). Let $d_{Z}$ be the dimension of latent space $Z$. We consider the case of a binary classification problem. 
We employ a model composed of a linear classifier over the features provided by $f_{I}$. The linear layer providing the logits of the $2$ different classes is defined as $Wz$ where $z \in Z$ and $W$ is the $2\times d_{Z}$ weight matrix.
More formally, the  linear classifier probe model outputs $y_{pred}$ as:
\begin{equation}
\label{eq:linear_probe}
y_{pred} = \textup{argmax}_{k\in\{0,1\}} Wf_{I}(x)
\end{equation}
If $w_{0*}$, $w_{1*}$ are the two rows of matrix $W$, then $y_{pred}$ can be written as 
\begin{equation}\label{eq:hyperplane}
y_{pred} = 
\begin{cases} 
    1, & \text{if } \vec{n} \cdot f_{I}(x) \geq 0, \\ 
    0, & \text{if } \vec{n} \cdot f_{I}(x) < 0
\end{cases}
\end{equation}
where the normal
\begin{equation}\label{eq:normal}
    \vec{n} = \frac{w_{1*}-w_{0*}}{\|w_{1*}-w_{0*}\|}
\end{equation}
is unitary and perpendicular to the hyperplane separating embeddings predicted as class $\bf{0}$ (negative) from those predicted as class $\bf{1}$ (positive).
Lastly, we use $f_{T}$ to refer to the text encoder sharing the same latent space as image encoder $f_{I}$.

\subsection{Create the concept-based dataset}

To avoid relying on a predefined set of bias-inducing candidate concepts, we create a concept dataset for use in the CUBIC methodology across any classification task. To construct this concept dataset, we extract name phrases from a text corpus, specifically the descriptions provided in the Conceptual Captions dataset \cite{sharma2018conceptual}. Name phrases are sequences of words that typically include nouns, adjectives, and articles, representing meaningful concepts within a caption. For instance, given the caption "a bird flying over a water tank", we extract the concepts bird, water, tank, and water tank. Following the extraction process, we perform deduplication to remove repeated concepts, resulting in a final set of approximately 160k unique concepts.

The choice to extract concepts from a caption dataset, rather than from a dictionary, is motivated by the fact that the CLIP backbone \cite{radford2021learning} is trained on image-caption pairs. This means that the representations in the semantic space will capture the semantics of concepts grounded in visible contexts. The text corpus provided by the descriptions in the Conceptual Captions dataset \cite{sharma2018conceptual}, with its vast number and diversity of described scenarios, serves effectively as a diverse and broad source of visual concepts. This diversity allows CUBIC to detect a wide range of fine-grained concepts, unconstrained by the limited vocabulary of captioning methods \cite{wang2019describing}.

\subsection{Compute the CUBIC bias score}
This metric measures the bias induced on a linear classifier probe by a given concept. Let $C$ be that concept, $f_{T}$ the text encoder of the VLM, and $\vec{n}$ the vector normal to the hyperplane as defined in Eq. \eqref{eq:normal}. 
Given a superclass label $\mathcal{L}$ shared by all images in the dataset, we define the CUBIC bias score $\cos{\alpha_{C}}$, which takes values in the range $[-1, 1]$, as
\begin{equation}\label{eq:simscore}
    \cos{\alpha_{C}} = \vec{n} \cdot \frac{f_{T}(\mathcal{L} \:\&\: C) - f_{T}(\mathcal{L})}{\|f_{T}(\mathcal{L} \:\&\: C) - f_{T}(\mathcal{L})\|},
\end{equation}
where $\mathcal{L} \:\&\: C$ represents the text concept $C$ combined together with the superclass label $\mathcal{L}$, which is shared by all images in the dataset. We observe that the term $f_{T}(\mathcal{L}\:\&\: C) - f_{T}(\mathcal{L})$ in the right-hand-side of Eq. \eqref{eq:simscore} captures the concept-driven shift of the superclass $\mathcal{L}$ embedding. If $\cos\alpha_{C} > 0$, the concept-driven shift of the superclass embedding $f_{T}(\mathcal{L} \:\&\: C) - f_{T}(\mathcal{L})$ points toward the class $\bf{1}$ side of the hyperplane. This means that when the concept $C$ is combined with superclass label $\mathcal{L}$, it \textit{pushes} the superclass embedding in the direction of the normal, contributing to the model predicting class $\bf{1}$. 

We decide to combine $\mathcal{L}$ and $C$ into $\mathcal{L} \:\&\: C$ textually as comma-separated concepts: $C, \mathcal{L}$. For example, in the CelebA \cite{liu2015faceattributes} dataset, which contains images of famous people with various annotated attributes that can be employed as target, we use $\mathcal{L}=person$, so for the concept $C=Eyeglasses$ we have $\mathcal{L}\:\&\: C = person,\;Eyeglasses$.
We highlight the geometrical meaning of $\cos{\alpha_{C}}$ defined in Eq. \eqref{eq:simscore}, which is equal to the cosine of the angle $\alpha_{C}$ between the normal $\vec{n}$ to the separation hyperplane 
 and the vector $f_{T}(\mathcal{L}\:\&\: C) - f_{T}(\mathcal{L})$, illustrated in  ~\Cref{fig:angle_between_concept_and_hyperplane}.

\noindent
\subsection{Identify bias-inducing concepts}
After computing the CUBIC score $\cos{\alpha_{C}}$ for all concepts, those with values closest to $1$ or $-1$ are the ones the classifier most strongly associates with classes $\bf{1}$ and $\bf{0}$, respectively. However, these concepts may include both bias-inducing concepts and legitimate predictive features expected to contribute to correct classification.
For example, in the task of classifying images with eyeglasses, where the dataset contains most glasses-wearers being blonde, the top concepts indicated by the higher absolute values of $\cos{\alpha_{C}}$ might include both valid predictors ('sunglasses', 'spectacles') and potentially biasing concepts ('blonde hair'). To identify bias-causing concepts, it is crucial to remove the concepts that are valid for prediction. Filtering out non-biasing concepts, as in the previous example, can be efficiently achieved through a programmatic method. Our approach leverages the BART-Large-MNLI \cite{lewis2019bart} zero-shot classifier to automatically determine whether a concept is bias-related. For instance, in the glasses detection example, the classifier labels concepts as either glasses-related or non-glasses-related, allowing us to retain only the latter as bias-inducing concepts. 

\section{Experimental Settings}\label{sec:experiments}

This section introduces the datasets used for our experiment, the selective undersampling procedure used to control different degrees of concept-induced biases, training details, and evaluation metrics. Datasets used contain concept annotations, which we will use to validate the effectiveness of CUBIC. However, it is important to note that our methodology works without requiring such annotations.

\subsection{Datasets}\label{subsec:datasets}
For our experiments, we use Waterbirds \cite{sagawadistributionally}, a dataset of waterbirds and landbirds with labeled water and land background environments, and CelebA \cite{liu2015faceattributes}, an extensive dataset of face attributes. Since the CelebA is a dataset of faces of famous people containing annotation of 40 different attributes which could be used as class targets for the classification problem, we will denote CelebA-Hat when the target chosen is whether the person wears or not hat, CelebA-Smile when the target is to determine whether the person is or not smiling, etc. As there are 40 annotated attributes on the CelebA dataset, we keep only half of them, giving preference to those attributes representing the entire presence or absence of an objective concept and not indicating subjective concepts or related to size/scale, etc., e.g., we include \textit{wears hat?} but not \textit{is beautiful?} or \textit{has big nose?}. 

Rather than relying on the original splits, we implement a selective undersampling procedure similar to that employed by \cite{jain2023distilling} to control different degrees of concept-induced biases. In particular, for a given concept \( C \) and class \( k \), we measure the Class-Concept Dataset Disagreement Ratio, which is defined as
\begin{equation}\label{eq:defintheta}
     \theta = \frac{\left|\{(x,y):(y=k\land \neg C)\lor (y\neq k \land C)\}\right|}{\left|\{(x,y):(y=k\land C)\lor (y\neq k \land \neg C)\}\right|} 
 \end{equation}

The denominator counts all images where $C$ and $k$ co-occur or are both absent, while the numerator counts all images where either $C$ or class $k$ occurs, but not both simultaneously.
 For fixed concept \( C \) and class \( k \), we choose a $\theta$ and generate a dataset by undersampling. E.g., if we choose $\theta=0$ to generate a dataset that complies with $\theta$, this means according to Eq. \eqref{eq:defintheta} that concept $C$ will be present in all images with class $k$ and absent from all images with ground-truth class different from $k$. After undersampling a dataset for a given $\theta\in\{0,0.05, 0.1,0.15,0.2,0.4\}$, we create splits for training, validation, and test from it.

 For CelebA datasets (CelebA-Hat, CelebA-Eyeglasses, etc.), the target class and the concept $C$ are different attributes chosen from a list of 20 annotated attributes in the original CelebA. 
 In the Waterbirds dataset, classes are always waterbird and landbird, and the concept will always be the background type.
 For each target-concept pair, 12 biased dataset are created by undersampling with $\theta\in\{0,0.05, 0.1,0.15,0.2,0.4\}$ (see Table \ref{table:ratios_waterbirds_main}). In total, we have 12 undersampled datasets for the Waterbirds case and more than 4000 for CelebA.

\subsection{Training details.}
We use as $f_{I}$ the CLIP \cite{radford2021learning} ViT B-32 image encoder, a visual transformer that provides embeddings in a 512-dimensional latent space ($d_{Z}=512$). We keep $f_{I}$ frozen and train only the parameters of $W$ from Eq. \eqref{eq:linear_probe} with a cross-entropy loss on the training dataset of the classification task. Each linear classifier probe is trained with the standard cross-entropy loss for a maximum of 200 epochs with warmups iterations and early stopping. We employ the Stochastic Gradient Descent (SGD)  optimizer with an initial learning rate of \(10^{-3}\) and momentum set to 0.9.
Since only the last linear layer is trainable, we store the latent representations of the images to avoid recomputing the features from the CLIP extractor during each forward pass. 
All experiments have been performed using a single Titan RTX GPU. Training for a model takes less than 20 secs.

\subsection{Metrics}

\noindent \textbf{Ground Truth Model Bias $\Delta_{C}$.}
Given a linear classifier probe, we need a metric to determine whether concept $C$ influences the classification output. 
We define:  
\begin{equation}
    TR_{k, C} = \frac{|(x,y) \in \mathcal{D} \;:\; (y = k) \land (y_{\text{pred}} = k) \land C|}{|(x,y) \in \mathcal{D} \;:\; (y = k) \land C|}
\end{equation}
Since \( TR_{1,C} \) represents the true positive rate for the subset of images where concept \( C \) is present, we denote it as \( TPR_C \). Similarly, \(TR_{0,C} \) corresponds to the true negative rate, which we denote as \( TNR_C \).  
Analogously, we define \( TPR_{\neg C} \) and \( TNR_{\neg C} \) for the group of images where concept \( C \) is absent.

 We define \begin{equation}\label{eq:delta}
    \Delta_{C} = 100\cdot\frac{(\textup{TPR}_{C} - \textup{TPR}_{\neg C}) - (\textup{TNR}_{C} - \textup{TNR}_{\neg C})}{2}
\end{equation}
where $\Delta_{C}\in [-100,100]$ is a measure of ground-truth model bias induced by concept $C$. It represents how much more sensitive class $\bf{1}$ is to concept $C$ than class $\bf{0}$.  When \(\Delta_{C}\) is close to \(100\), it indicates that the model interprets the presence of concept \(C\) as very strong evidence in favour of class $\bf{1}$. Conversely, if \(\Delta_{C}\) is close to \(-100\), the model considers the presence of \(C\) as strong evidence in favour of class $\bf{0}$. Finally, \(\Delta_{C} \approx 0\) implies that the presence of \(C\) does not influence the model’s preference towards one class over the other. 

\noindent
\textbf{Spearman correlation}. It represents the correlation between two variables \( A \) and \( B \), denoted as \(\text{Spearman}(A, B)\), and it measures the strength and direction of their monotonic relationship. It ranges from $-1$ to $1$, where $-1$ indicates a perfect negative monotonic relationship (as \( A \) increases, \( B \) decreases), $1$ indicates a perfect positive monotonic relationship (both increase together), and $0$ signifies no monotonic association. 

\noindent
\textbf{TopK Bias Detection Accuracy}. Given a set of bias concept candidates, we denote as $C_{\textup{class $\bf{1}$}}$ the concept with higher $\Delta_{C}$  and as $C_{\textup{class $\bf{0}$}}$ the concept with lowest $\Delta_{C}$. Given the top \( K \) predictions of concepts biasing the model toward class $\bf{1}$, we check if any of them matches \( C_{\textup{class $\bf{1}$}} \). The percentage of times this occurs defines the TopK Bias Detection accuracy for class $\bf{1}$ bias prediction. The same procedure is applied to concepts biasing the model toward class $\bf{0}$. We then average the TopK accuracies for both classes and refer to this as the TopK bias detection accuracy.

\section{Evaluation}\label{sec:evaluation}
This evaluation section consists of two main stages. First, we directly compare CUBIC to established bias detection methods. This allows us to benchmark its effectiveness and identify any notable discrepancies. Next, we explore key research questions concerning the behavior of the CUBIC score, the relationship between dataset and classifier bias, the identification of finer-grained concepts, the use of superclass labels and the influence of concept dataset sources.

\def\arraystretch{1}
\begin{table}[htbp]
\centering
\caption{Top 1, Top 3 and Top 5 bias detection accuracies. \textit{No Cand.} refers to the method not requiring the use of a predefined set of concept candidates to test bias on  the experiment.}
\begin{tabular}{p{0.7cm}p{0.7cm}lcccc} 
\toprule 
No Cand. & Data & Method & 
\(\textup{Top1 Acc.}\) & \(\textup{Top3 Acc.}\) & \(\textup{Top5 Acc.}\) \\ 
\midrule  
\textcolor{red}{\ding{55}} &\multirow{2}{*}{Waterb.} & DrML\cite{zhang2023diagnosing} & 100\% & - & -\\
\textcolor{red}{\ding{55}} && CUBIC  & 100\% & - & -\\
\midrule  
\textcolor{red}{\ding{55}}&\multirow{2}{*}{CelebA} & DrML\cite{zhang2023diagnosing} & 27.17\% & 49.44\% & 67.01\% \\
\textcolor{red}{\ding{55}}&& CUBIC  & \textbf{31.05\%} & \textbf{57.75\%} & \textbf{73.07\%} \\
\midrule  
\midrule
\textcolor{darkgreen}\checkmark&\multirow{2}{*}{Waterb.} & B2T\cite{kim2024keyword} & \textbf{33.33\%} & 54.17\% & 62.50\\
\textcolor{darkgreen}\checkmark&& CUBIC  & 29.17\% & \textbf{58.33\%} & \textbf{70.83\%} \\
\midrule  
\textcolor{darkgreen}\checkmark&\multirow{2}{*}{CelebA} & B2T\cite{kim2024keyword} & 4.39\% & 9.26\% & 14.46\% \\
\textcolor{darkgreen}\checkmark&& CUBIC  & \textbf{21.25\%} & \textbf{32.38\%} & \textbf{37.31\%} \\
\bottomrule 
\end{tabular}
\label{table:topK_accuracy}
\vspace{-10pt}
\end{table}

\subsection{Main comparisons with bias discovery methods.}

In~\Cref{table:topK_accuracy} CUBIC is compared with the most relevant baselines in concept-bias discovery, i.e., DrML\cite{zhang2023diagnosing} and B2T\cite{kim2024keyword} on CelebA and Waterbirds datasets \cite{zhang2023diagnosing}.
In the setting where methods predict bias from a predefined set of candidate concepts, the Waterbirds dataset is evaluated only for Top1 accuracy. This is because it only contains two bias candidates, \textit{water background} and \textit{forest background}, meaning the top 2 predictions will always include the correct answer.
From the table, results show that CUBIC performs better in detecting bias-concepts from a predefined set of candidates. In particular, it surpasses \textit{DrML} achieving \(+4.12\%\), \(+8.31\%\), and \(+6.06\%\) improvements for Top1, Top3, and Top5 bias detection accuracy, respectively.  
In addtion, when the list of candidates is unavailable, CUBIC produces similar or better results with respect to \textit{B2T}. In particular, for the CelebA case, CUBIC achieves improvements of +16.86\%, +23.12\%, and +22.85\% over B2T.
\subsection{Further analyses}

\begin{figure}[htbp]
    \centering
  \includegraphics[width=\linewidth]{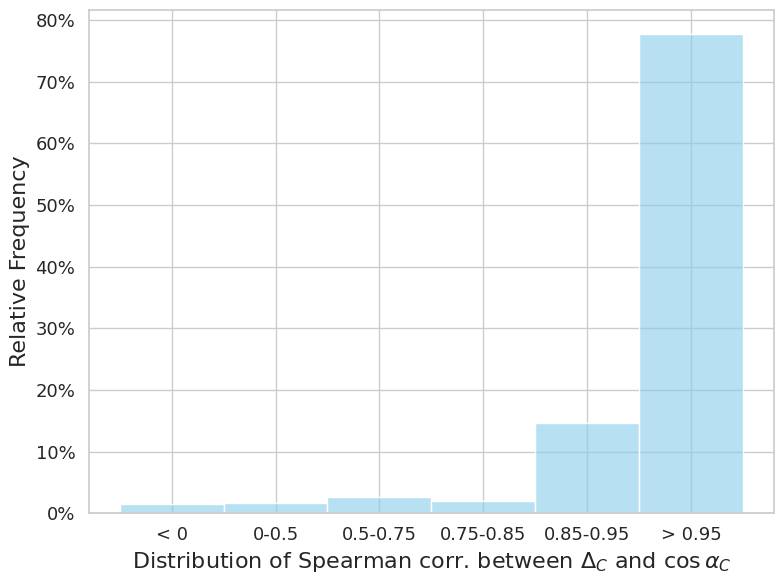}
    \caption{Distribution of the Spearman correlation between ground truth concept bias $\Delta_{C}$ and CUBIC scores ($\cos{\alpha_{C}}$) across multiple CelebA-derived datasets with $\theta$-controlled undersampling. The Spearman correlation distributions demonstrate the strong predictive power of $\cos{\alpha_{C}}$ in capturing concept-induced bias variations. Our results show that CUBIC scores achieve $>0.95$ spearman correlation with $\Delta_{C}$ almost 80\% of cases, validating its effectiveness as bias indicators.
    \vspace{-17pt}
    }
\label{fig:spearmandistribceleba}
\end{figure}

\setlength{\tabcolsep}{3pt}

\begin{table}[b]
\vspace{-10pt}
\caption{Ground-truth bias metric $\Delta_{C}$  and CUBIC 
$\cos{\alpha_{C}}$ bias score for classifiers trained via undersampling Waterbirds dataset \cite{sagawadistributionally}. "$k$" denotes the class positively correlated with $C=$ Water background.
}
    \centering
    \begin{tabular}{p{0.5cm}p{1cm}ccccccc}
    \toprule
    \multicolumn{8}{c}{Concept $C=\:$Water background} \\ 
    \midrule
    $k$ & \textbf{Metrics} & $\theta$=0 & $\theta$=0.05 & $\theta$=0.1 & $\theta$=0.15 & $\theta$=0.2 & $\theta$=0.4 & $\theta=1$\\
    \midrule
 \multirow{2}{*}{1} & $\Delta_{C}$ & 74.099 & 49.276 & 53.348 & 42.662 & 37.829 & 20.614 & 2.097 \\
 & $\cos{\alpha_{C}}$  & 0.133 & 0.129 & 0.128 & 0.127 & 0.116 & 0.084 & 0.045 \\
\midrule
  \multirow{2}{*}{0} & $\Delta_{C}$ & -72.34 & -58.271 & -47.256 & -36.43 & -29.747 & -11.4 & 2.665 \\
 & $\cos{\alpha_{C}}$  & -0.084 & -0.065 & -0.046 & -0.036 & -0.028 & -0.002 & 0.033 \\
 \midrule
 \midrule
     \multicolumn{8}{c}{Spearman$(\Delta_{C}$,$\cos{\alpha_{C}}) = 0.99$}\\
    \bottomrule
    \end{tabular}
\label{table:ratios_waterbirds_main}

\end{table}

\textbf{Does CUBIC bias score increase as the classifier bias increases?}
To detect non-linear correlation, \Cref{fig:spearmandistribceleba} shows Spearman correlation between the ground truth bias and the CUBIC score. In particular, we aim to test that, for a given concept $C$, CUBIC score is monotonically increasing with ground truth bias $\Delta_{C}$. From the experiment, it is visible that such correlation allows the CUBIC score to be used to compare two models and find which one is more biased towards a particular concept. This is shown also in  ~\Cref{table:ratios_waterbirds_main}. Here, this table presents the values of \( \Delta_{C} \) along with the \( \cos{\alpha_{C}} \) scores for linear classifier probes trained on 12 undersampled Waterbirds data. The Spearman correlation between the ground-truth model bias metric \( \Delta_{C} \) and the CUBIC score \( \cos{\alpha_{C}} \) is 0.99, indicating that the CUBIC score effectively reflects changes in bias.

On the other hand, \Cref{fig:spearmandistribceleba} shows the distribution of the Spearman correlation coefficient, \(\text{Spearman}(\Delta_{C}, \cos{\alpha_{C}})\), computed for models trained on the undersampled CelebA dataset. The vast majority of Spearman coefficients are high, indicating that, in most cases, the CUBIC score \( \cos{\alpha_{C}} \) exhibits a strong monotonic relationship with the ground-truth model bias. Therefore, given two linear classifier probes trained on the same task, the CUBIC score can determine which classifier is more biased by a specific concept \( C \).

\textbf{Does CUBIC detect bias towards finer-grained concepts?}

\begin{figure}[b]
\vspace{-20pt}
\small
\begin{tabular}{lccc}
\toprule
Coarse & \multicolumn{3}{c}{Fine-grained concepts}  \\
\midrule
Forest & 
\begin{minipage}{0.265\linewidth}
    \centering
    \includegraphics[width=0.9\linewidth,height=0.9\linewidth]{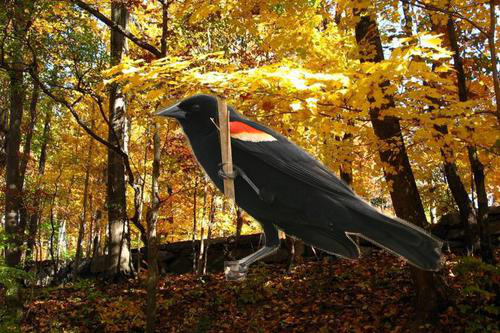} \\
    Yellow Forest $\cos{\alpha_{C}}=-0.190$
\end{minipage} & 
\begin{minipage}{0.265\linewidth}
    \centering
    \includegraphics[width=0.9\linewidth,height=0.9\linewidth]{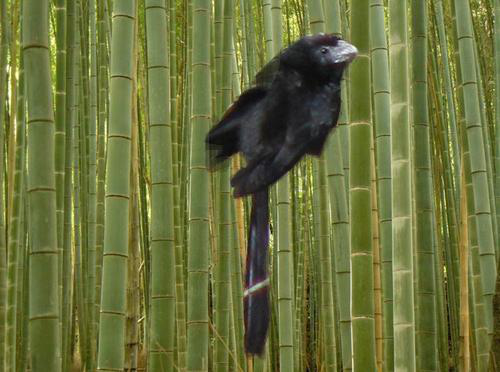} \\
    Bamboo Forest
    $\cos{\alpha_{C}}=-0.175$
\end{minipage} & 
\begin{minipage}{0.265\linewidth}
    \centering
    \includegraphics[width=0.9\linewidth,height=0.9\linewidth]{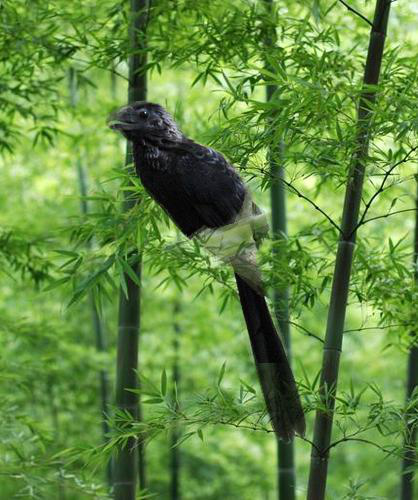} \\
    Green Canopy
    $\cos{\alpha_{C}}=-0.172$
\end{minipage} \\
\midrule
Water & 
\begin{minipage}{0.265\linewidth}
    \centering
    \includegraphics[width=0.9\linewidth,height=0.9\linewidth]{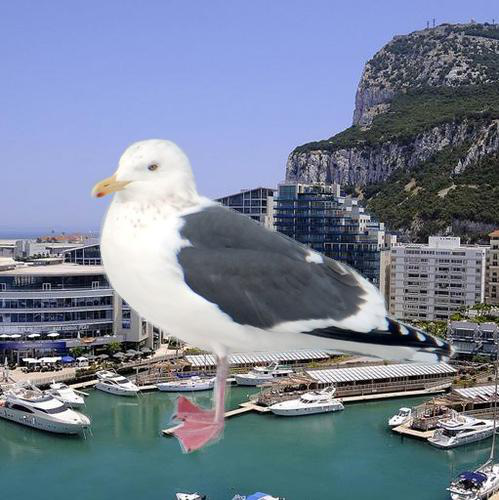} \\
    Presidential Yacht
    $\cos{\alpha_{C}}=0.164$
\end{minipage} & 
\begin{minipage}{0.265\linewidth}
    \centering
    \includegraphics[width=0.9\linewidth,height=0.9\linewidth]{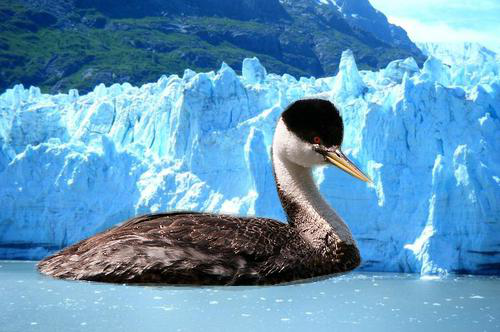} \\
    Iceberg Lake
    $\cos{\alpha_{C}}=0.155$
\end{minipage} & 
\begin{minipage}{0.265\linewidth}
    \centering
    \includegraphics[width=0.9\linewidth,height=0.9\linewidth]{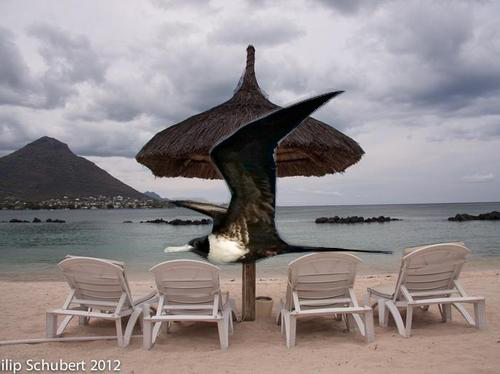} \\
    Cloudy Beach
    $\cos{\alpha_{C}}=0.136$
\end{minipage} \\
\bottomrule
\end{tabular}
\caption{Images retrieved from the Waterbirds dataset \cite{sagawadistributionally} evidencing the most influential biasing concepts discovered by CUBIC. CUBIC identifies finer-grained concepts beyond just \textit{forest background} and \textit{water background}.}
\label{fig:top_cubic_predictions}
\end{figure}

\noindent  
~\Cref{fig:top_cubic_predictions} provides qualitative evidence that the CUBIC score can detect not only coarse-grained bias concepts, such as \textit{forest}, but also fine-grained ones, including \textit{yellow forest}, \textit{bamboo forest}, and \textit{green canopy}. 

\textbf{Does dataset bias always imply linear probe classifier bias?
}
A lower Class-Concept Dataset Disagreement Ratio \(\theta\) signifies a stronger link between class \(k\) and concept \(C\), leading the classifier to associate concept \(C\) with class \(k\). This trend is generally observed in \Cref{table:ratios_waterbirds_main}. However, when class \(k=1\) correlates with the Water background at \(\theta\) values of 0.05 and 0.1, an unexpected pattern emerges. While \(|\Delta_{C}|\) for \(\theta=0.1\) should be lower than for \(\theta=0.05\) (indicating weaker bias due to a looser feature connection), this is not the case. This anomaly can arise from factors such as training randomness, variations in the dataset or labels, and the feature representations learned by the backbone network.

\begin{table}[!t]
    \caption{Alternatives for the second term of $\cos \alpha_C$ in Eq.~\eqref{eq:simscore} (before dividing by its Euclidean norm). Accuracy in CelebA undersampled datasets.
}
    \centering
    \begin{tabular}{cc}
    \toprule
        Concept embedding factor & Acc. \\
        \midrule 
        $f_{T}(C)$ & 18.21\% \\
        $f_{T}(C \: \& \: \mathcal{L}) - f_{T}(\mathcal{L})$ & 20.58\% \\
        $ \boldsymbol{f_{T}(\mathcal{L} \: \& \: C) - f_{T}(\mathcal{L})}$ & \textbf{21.25\%}\\
        \bottomrule
    \end{tabular}
\vspace{-20pt}\label{tab:superclasslabelalternatives}
\end{table}

\noindent
\textbf{Do we require a superclass label?}
 In ~\Cref{tab:superclasslabelalternatives}, we present an ablation study exploring alternative definitions for \(\cos{\alpha_{C}}\) beyond the one provided in Eq.~\eqref{eq:simscore}. Our results confirm that utilizing a superclass label to compute the CUBIC score, especially when given before the concept to be tested, yields the highest accuracy in predicting bias within the CelebA undersampled dataset.
 The superclass label can help to resolve ambiguity by explicitly specifying the intended sense of a concept, ensuring that the model interprets it correctly. Without this clarification, a text encoder can misinterpret a concept with multiple meanings, leading to unintended errors in bias prediction. By using a context superclass label, we make sure the embedding produced by the text encoder is that of the intended concept meaning.

\noindent
\textbf{How relevant is the choice of the concept source?}
~\Cref{tab:conceptsources} presents accuracy results on the CelebA dataset for the CUBIC methodology, applied using different sources for concept dataset creation. The results confirm that visually grounded sources, such as Conceptual Captions \cite{sharma2018conceptual} or COCO captions \cite{lin2014microsoft}, lead to better performance. In contrast, concepts extracted from a knowledge base (e.g.,  WordNet \cite{miller1995wordnet}), containing a proportion of non-visual concepts, yield worse results.

\begin{table}[!t]
    \caption{Possible sources for the Concept Dataset.}
    \centering
    \begin{tabular}{cc}
    \toprule
        Concept Dataset Source & Accuracy\\ 
        \midrule
        WordNet \cite{miller1995wordnet} & 16.99\% \\
        COCO captions \cite{lin2014microsoft} & 20.50\% \\
        \textbf{Conceptual captions} \cite{sharma2018conceptual} & \textbf{21.25\%} \\
        \bottomrule
    \end{tabular}
    \vspace{-12pt}
    \label{tab:conceptsources}
\end{table}

\section{Discussion}
\label{sec:discussion}
\noindent
In this work, we introduced the CUBIC methodology for identifying and addressing bias in both scenarios: one where bias must be detected from a predefined set of candidate concepts, and another where bias must be determined without prior knowledge of which concepts are prone to induce it.
 CUBIC does not require specific underperforming image samples in the training, validation, or test sets.

\noindent
\textbf{CUBIC applications}. CUBIC can discover concept-based biases, facilitating actionability through bias mitigation methods such as dataset modification, robust optimization, or retraining with text data \cite{zhang2023diagnosing}. Dataset modification includes undersampling overrepresented groups or augmenting underrepresented groups with real or synthetic data \cite{kumar2024image}. Distributionally robust optimization (DRO) \cite{sagawadistributionally} ensures equitable performance by minimizing worst-case errors across subpopulations. 

\noindent
\textbf{Inherited limitations}.
\noindent  
It is important to note that CUBIC inherits the limitations of CLIP's feature representations. Since our method fine-tunes a linear classifier on top of CLIP's vision backbone, it is unlikely to generate concepts whose visual or semantic features are not effectively captured by CLIP's image and text encoders.
This limitation impacts specialized fields like medical imaging. Just as bias detection relying on captioning methods must ensure captioners are adapted to the specialized domain, our approach requires additional considerations.
Besides fine-tuning the VLM backbone (necessary for classification purposes), we must ensure specialized concepts are represented in our concept dataset. Finally, we note that any improvements in the representations generated by VLM encoders will directly enhance CUBIC's effectiveness.

\vspace{-2pt}

\section{Conclusion and Future Works}\label{sec:conclusion}
\noindent
Our experiments demonstrate that CUBIC provides a novel and powerful approach to bias identification. Unlike traditional bias detection techniques that rely on performance disparities across subgroups, CUBIC can identify bias in a linear classifier probe without requiring access to failure cases linked to the bias. 
In terms of future work, extending the CUBIC method to a multiclass setting and verifying its effectiveness on different backbones and datasets are natural extensions, enabling a broader applicability of the method. Besides, finding a precise way to estimate ground truth bias $\Delta_{C}$ as a function of the CUBIC score $\cos{\alpha_{C}}$ remains an open challenge to quantify how critical the bias is.
Finally, we emphasize that detecting bias induced by concepts stemming from spurious correlations in training data is crucial for preventing models from relying on sensitive attributes, hence promoting fairness and transparency in AI. Within transparency as a trustworthy AI requirement \cite{diaz2023connecting}, explainability is paramount in building trust in models.

\section{Acknowledgements}
Work supported by Arqus Talent Scholarship, the 2022 Leonardo Grant (BBVA foundation) and the TSI-100927-2023-1 Project (Transformation and Resilience Plan from the EU NextGen through the Ministry for Digital Transformation and the Civil Service).  Confalonieri acknowledges ‘NeuroXAI’ (BIRD231830) funding.

\bibliographystyle{IEEEtran}
\bibliography{references_cleaned}

\end{document}